\documentclass[runningheads]{llncs}
\usepackage{float}
\usepackage[T1]{fontenc}
\usepackage{graphicx,verbatim}
\usepackage{subcaption}
\usepackage{booktabs}
\usepackage{tabularx}
\usepackage{subfig}
\usepackage{amsmath,amssymb}
\begin{document}
\setlength{\floatsep}{5pt plus 2pt minus 2pt}     
\setlength{\textfloatsep}{5pt plus 2pt minus 2pt} 
\setlength{\intextsep}{5pt plus 2pt minus 2pt}    
\setlength{\abovedisplayskip}{1pt} 
\setlength{\belowdisplayskip}{1pt} 
\setlength{\abovedisplayshortskip}{0pt} 
\setlength{\belowdisplayshortskip}{0pt} 

\title{MO-RiskVAE: A Multi-Omics Variational Autoencoder for Survival Risk Modeling in Multiple MyelomaMO-RiskVAE}
%
\author{
Zixuan Chen$^{1}$\thanks{These authors contributed equally to this work.} \and
Heng Zhang$^{1}$\protect\footnotemark[1] \and
YuPeng Qin$^{1}$ \and
WenPeng Xing$^{1}$ \and
Qiang Wang$^{2}$ \and
Da Wang$^{3}$ \and
Changting Lin$^{1}$ \and
Meng Han$^{1}$\thanks{Corresponding author.}
}

\authorrunning{Chen et al.}

\institute{
Binjiang Institute of Zhejiang University \\
\email{annie\_021104@163.com, HZHANG86@e.ntu.edu.sg, qinyupeng177@gmail.com, 843553050@qq.com, linchangting@gmail.com, mhan@zju.edu.cn}
\and
Zhejiang University \\
\email{wangqiang1983@zju.edu.cn}
\and
School of Medicine, Zhejiang University \\
\email{wangdao618@zju.edu.cn}
}
  
\maketitle

\begin{abstract}
Multimodal variational autoencoders (VAEs) have emerged as a powerful framework
for survival risk modeling in multiple myeloma by integrating heterogeneous
omics and clinical data. However, when trained under survival supervision,
standard latent regularization strategies often fail to preserve
prognostically relevant variation, leading to unstable or overly constrained
representations. Despite numerous proposed variants, it remains unclear which
aspects of latent design fundamentally govern performance in this setting.
In this work, we conduct a controlled investigation of latent modeling choices
for multimodal survival prediction within a unified extension of the MyeVAE
framework. By systematically isolating regularization scale, posterior geometry,
and latent space structure under identical architectures and optimization
protocols, we show that survival-driven training is primarily sensitive to the
magnitude and structure of latent regularization rather than the specific
divergence formulation. In particular, moderate relaxation of KL regularization
consistently improves survival discrimination, while alternative divergence
mechanisms such as MMD and HSIC provide limited benefit without appropriate
scaling.
We further demonstrate that structuring the latent space can improve alignment
between learned representations and survival risk gradients. A hybrid
continuous--discrete formulation based on Gumbel--Softmax enhances global risk
ordering in the continuous latent subspace, even though stable discrete subtype
discovery does not emerge under survival supervision. Guided by these findings,
we instantiate a robust multimodal survival model, termed MO-RiskVAE, which
consistently improves risk stratification over the original MyeVAE without
introducing additional supervision or complex training heuristics.


\end{abstract}
\section{Introduction}

Multiple myeloma (MM) exhibits substantial clinical and genetic heterogeneity,
leading to highly variable survival outcomes across patients
\cite{chang2025myevae}. Accurate risk stratification at diagnosis is therefore
critical for treatment planning and prognosis assessment. However, conventional
clinical scoring systems, such as the Revised International Staging System
(R-ISS), remain limited in capturing the molecular complexity underlying disease
progression, motivating data-driven survival models based on high-dimensional
omics profiling.

Multimodal representation learning provides a principled framework for
integrating heterogeneous omics data, including transcriptomic, genomic, and
cytogenetic modalities, to improve survival prediction
\cite{shi2019moe,hira2021integrated}. Variational autoencoders (VAEs) are
particularly attractive in this context due to their ability to learn compact
latent representations from high-dimensional inputs while naturally handling
missing modalities \cite{hira2021integrated,qiu2024barycentric}. In MM, the
MyeVAE framework \cite{chang2025myevae} further couples multimodal latent learning
with a Cox proportional hazards objective, enabling end-to-end optimization
directly driven by survival outcomes.

Despite their empirical success, VAEs trained under survival supervision exhibit
failure modes that remain poorly understood. In contrast to reconstruction- or
likelihood-driven settings, survival objectives such as the Cox loss dominate
the optimization landscape and fundamentally alter how latent regularization
shapes the learned representation. Standard evidence lower bound (ELBO)
formulations impose sample-wise Kullback--Leibler (KL) constraints that, when
combined with strong survival gradients, can suppress prognostically relevant
variation and lead to unstable or overly constrained latent spaces
\cite{hassan2025mmdvae}. While alternative regularization strategies, including
$\beta$-reweighted KL, maximum mean discrepancy (MMD), and dependence-based
penalties such as HSIC, have been explored, it remains unclear whether their
effectiveness arises from their specific formulation or from the scale and
structure with which they interact with survival-driven training.

Beyond regularization strength, the structure of the latent space itself plays a
critical role in survival modeling. Disease progression in MM reflects both
continuous risk trajectories and discrete sources of biological heterogeneity.
Fully continuous latent representations may struggle to capture this dual
nature, while discrete or hybrid formulations, such as VQ-VAE
\cite{oord2017vqvae} and Gumbel--Softmax relaxations \cite{jang2016gumbel}, offer a
potential mechanism for incorporating latent heterogeneity. However, the
stability and practical benefit of such structured latent designs under
survival supervision, particularly in moderate-size multimodal cohorts, remain
largely unexplored.

In this work, we conduct a unified and controlled investigation of latent
regularization and latent structure for multimodal survival analysis within the
MyeVAE framework. By systematically isolating regularization scale, posterior
geometry, and latent space structure under identical architectures and training
protocols, we show that survival-driven optimization is governed primarily by
the scale and structure of latent regularization rather than the specific choice
of divergence. Guided by these findings, we instantiate a principled multimodal
survival model, termed \textbf{MO-RiskVAE}, which achieves consistently improved
risk stratification over the original MyeVAE without introducing additional
supervision or complex training heuristics.

\section{Method}

\noindent\textbf{Survival-Supervised Multimodal VAE Framework}.
We consider a multimodal variational autoencoder trained under survival
supervision, following the MyeVAE framework. Given multimodal input $x$, an
encoder $q_\phi(z|x)$ maps observations into a latent representation $z$, which
is reconstructed by a decoder $p_\theta(x|z)$. A survival prediction head
$f_\psi(z)$ is jointly optimized using a Cox proportional hazards objective.
In this setting, survival gradients often dominate optimization, motivating a
systematic examination of how latent regularization scale, geometry, and
structure interact with survival-driven training.

\noindent\textbf{Design Axis I: Regularization Scale via $\beta$-KL}.
We first isolate the effect of latent regularization \emph{scale} using a
$\beta$-reweighted Kullback--Leibler (KL) divergence. Rather than introducing a
new divergence, this formulation provides a minimal and continuous probe for
controlling posterior constraint strength under survival supervision.

To mitigate over-regularization from standard KL in MyeVAE, we adopt a
$\beta$-VAE that reweights the KL term with $\beta \in (0,1)$, increasing latent
flexibility while preserving the original ELBO structure. For input $x$, the
encoder $q_\phi(z|x)$ and decoder $p_\theta(x|z)$ optimize:
\begin{equation}
\label{eq:beta_vae}
\mathcal{L}_{\beta\text{-VAE}}(\theta, \phi; x) =
\mathbb{E}_{q_\phi(z \mid x)}[\log p_\theta(x \mid z)]
- \beta \, \mathrm{KL}(q_\phi(z \mid x) \| p(z)),
\end{equation}
where $p(z)$ is a standard normal prior. Setting $\beta < 1$ relaxes posterior
constraints and allows the latent space to retain prognostically relevant
variation that may otherwise be suppressed.

With a survival head $f_\psi(z)$, the end-to-end objective becomes:
\begin{equation}
\label{eq:joint_loss}
\mathcal{L}_{\mathrm{total}} =
\mathcal{L}_{\beta\text{-VAE}}
+ \lambda \, \mathbb{E}_{q_\phi(z \mid x)}
\big[ \mathcal{L}_{\mathrm{pred}}(f_\psi(z), y) \big],
\end{equation}
where $\lambda$ controls the strength of survival supervision.

\noindent\textbf{Design Axis II: Posterior Geometry and Dependence Control}.
Beyond regularization scale, we examine whether modifying posterior geometry or
explicitly suppressing spurious dependencies yields additional benefit under
survival-driven optimization.

\noindent\textbf{Distribution-Level Geometry Control (MMD).}
We replace the pointwise KL divergence with a distribution-level regularization
based on maximum mean discrepancy (MMD). While KL penalizes each conditional
posterior $q_\phi(z|x)$ independently, MMD aligns the aggregated posterior with
the prior, relaxing sample-wise constraints while preserving global
distributional structure.

The aggregated posterior is defined as
\begin{equation}
\label{eq:agg_posterior}
q(z) = \mathbb{E}_{x \sim p_\mathrm{data}(x)} [q_\phi(z|x)].
\end{equation}

MMD is computed as
\begin{equation}
\label{eq:mmd}
\begin{split}
\mathrm{MMD}(q(z),p(z)) =\;
& \mathbb{E}_{z,z'\sim q(z)}[k(z,z')] +
\mathbb{E}_{\tilde{z},\tilde{z}'\sim p(z)}[k(\tilde{z},\tilde{z}')] \\
& - 2 \mathbb{E}_{z \sim q(z), \tilde{z} \sim p(z)}[k(z,\tilde{z})],
\end{split}
\end{equation}
where $k(\cdot,\cdot)$ denotes an RBF kernel.

The overall objective is
\begin{equation}
\label{eq:mmd_vae_objective}
\mathcal{L}_\mathrm{MMD\text{-}VAE} =
\mathbb{E}_{q_\phi(z|x)}[\log p_\theta(x|z)]
- \lambda_\mathrm{MMD} \, \mathrm{MMD}(q(z),p(z))
+ \lambda_\mathrm{surv} \, \mathcal{L}_\mathrm{surv}.
\end{equation}

\noindent\textbf{Dependence-Aware Regularization (HSIC).}
To explicitly penalize spurious correlations induced by cohort, platform, or
modality-specific artifacts, we incorporate the Hilbert--Schmidt Independence
Criterion (HSIC) as a dependence-aware regularizer on prediction residuals.

Given residuals
\begin{equation}
r_i = y_i - f_\psi(z_i, c_i), \quad z_i \sim q_\phi(z|x_i),
\end{equation}
independence is measured by
\begin{equation}
\mathrm{HSIC}(X, r) = \frac{1}{(n-1)^2} \mathrm{tr}(KHLH),
\end{equation}
where $K$ and $L$ are kernel matrices and
$H = I - \frac{1}{n}\mathbf{1}\mathbf{1}^\top$.

The HSIC-augmented objective is
\begin{equation}
\begin{aligned}
\mathcal{L}_{\mathrm{HSIC}} =
&\; \mathbb{E}_{q_\phi(z \mid x)} [ \log p_\theta(x \mid z) ]
- \beta \, \mathrm{KL}(q_\phi(z \mid x) \| p(z)) \\
&+ \lambda \, \mathbb{E}_{q_\phi(z \mid x)}
[ \mathcal{L}_{\mathrm{pred}}(f_\psi(z, c), y) ] \\
&+ \gamma \, \mathrm{HSIC}\big(x, y - f_\psi(z, c)\big),
\end{aligned}
\end{equation}
where $\gamma$ controls the strength of dependence regularization.

\noindent\textbf{Design Axis III: Latent Space Structure via Gumbel--Softmax}.
In addition to regularization strength and geometry, latent space structure
constitutes a third critical design axis. We investigate a hybrid
continuous--discrete latent formulation using a differentiable Gumbel--Softmax
relaxation to examine whether explicit structural decomposition improves
alignment between latent representations and survival risk gradients.

\begin{figure}[H]
\centering
\includegraphics[width=\linewidth]{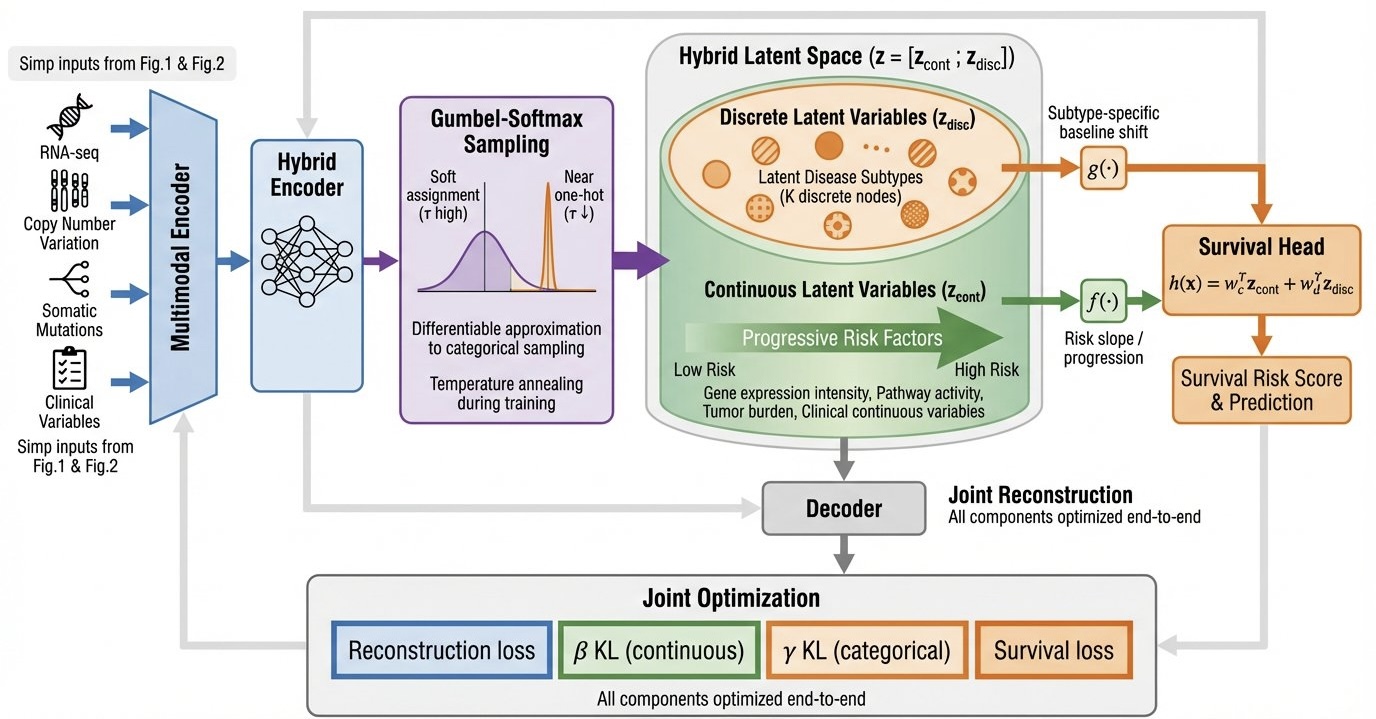}
\caption{Hybrid latent space formulation with Gumbel--Softmax.}
\label{fig:Figure3}
\end{figure}

The latent vector is decomposed as
\begin{equation}
z = [z_{\mathrm{cont}} ; z_{\mathrm{disc}}],
\end{equation}
where $z_{\mathrm{cont}}$ captures continuous disease variation and
$z_{\mathrm{disc}}$ encodes $K$ latent subtypes using Gumbel--Softmax:
\begin{equation}
y_k =
\frac{\exp((\log \pi_k + g_k)/\tau)}
{\sum_j \exp((\log \pi_j + g_j)/\tau)},
\quad g_k \sim \mathrm{Gumbel}(0,1),
\end{equation}
with temperature $\tau$ annealed toward one-hot assignments.

Survival risk is modeled as
\begin{equation}
h(x) = w_c^\top z_{\mathrm{cont}} + w_d^\top z_{\mathrm{disc}},
\end{equation}
explicitly separating continuous progression effects from subtype-specific
contributions.

The corresponding objective is
\begin{equation}
\mathcal{L} =
\mathcal{L}_{\mathrm{recon}}
+ \beta \, \mathrm{KL}(q_\phi(z_{\mathrm{cont}}|x)\|p(z))
+ \gamma \, \mathrm{KL}(q_\phi(z_{\mathrm{disc}}|x)\|\mathrm{Cat}(K))
+ \lambda \, \mathcal{L}_{\mathrm{survival}},
\end{equation}
which is compatible with $\beta$-, MMD-, or HSIC-based regularization.

\noindent\textbf{Final Instantiation: MO-RiskVAE}.Guided by the above controlled analysis, we instantiate a principled multimodal
survival model, termed \textbf{MO-RiskVAE}. The model adopts moderately relaxed
KL regularization to ensure an appropriate latent scale, incorporates structured
latent representations to improve global risk alignment, and avoids additional
regularization mechanisms that do not yield consistent benefit under survival
supervision. MO-RiskVAE therefore represents a concrete instantiation of the
design principles identified in this study, rather than an independent
architectural proposal.


\section{Experiments}

\noindent\textbf{Datasets and Experimental Protocol}.
All experiments follow the same datasets, preprocessing pipeline, and backbone
architecture as MyeVAE. The primary cohort is derived from the MMRF CoMMpass
study, consisting of 1,143 newly diagnosed multiple myeloma patients with six
modalities, among which 628 patients have complete multi-omics observations.
Four independent microarray cohorts (UAMS-GSE24080, HOVON-65/GMMG-HD4,
EMTAB-4032, and APEX) are used for external validation.

\begin{table}[H]
\centering
\small
\caption{Summary of data modalities used in this study.}
\label{tab:data_summary}
\begin{tabularx}{0.9\linewidth}{l X c c}
\toprule
Modality & Description & Samples & Features \\
\midrule
Clinical 
& Age, sex, ISS stage 
& 1,143 
& 3 \\
RNA-seq 
& Gene expression (log$_2$ TPM) 
& 767 
& High \\
CNV (WGS) 
& Copy number states ($-2$ to $+2$) 
& 877 
& High \\
SBS signatures (WES) 
& Mutational signature activities 
& 947 
& 10 \\
IgH translocation 
& Cytogenetic subtype labels 
& 783 
& 8 \\
Integrated 
& All modalities combined 
& 628 
& Composite \\
\bottomrule
\end{tabularx}

\footnotesize
Complete samples denote patients with all modalities available.
\end{table}

Modalities are aligned by patient identifier. Complete samples are split into
training and validation sets (1016/127), while partially observed samples are
retained for training. High-dimensional features are selected using
Elastic Net--regularized Cox models. Gene expression data are log-transformed
and standardized, CNV profiles are discretized into five states, and clinical
variables are normalized or one-hot encoded. Missing modalities are imputed
using an iterative round-robin strategy. External cohorts are processed using
identical normalization and feature mappings.

Unless otherwise specified, all models are trained as controlled extensions of
MyeVAE with a fixed latent dimension of 32, using Adam
($5\times10^{-4}$), batch size 1024, and early stopping over 100 epochs.
This shared protocol enables direct attribution of performance differences to
latent design choices rather than architectural or optimization confounders.

\begin{figure}[H]
\centering
\includegraphics[width=0.8\linewidth]{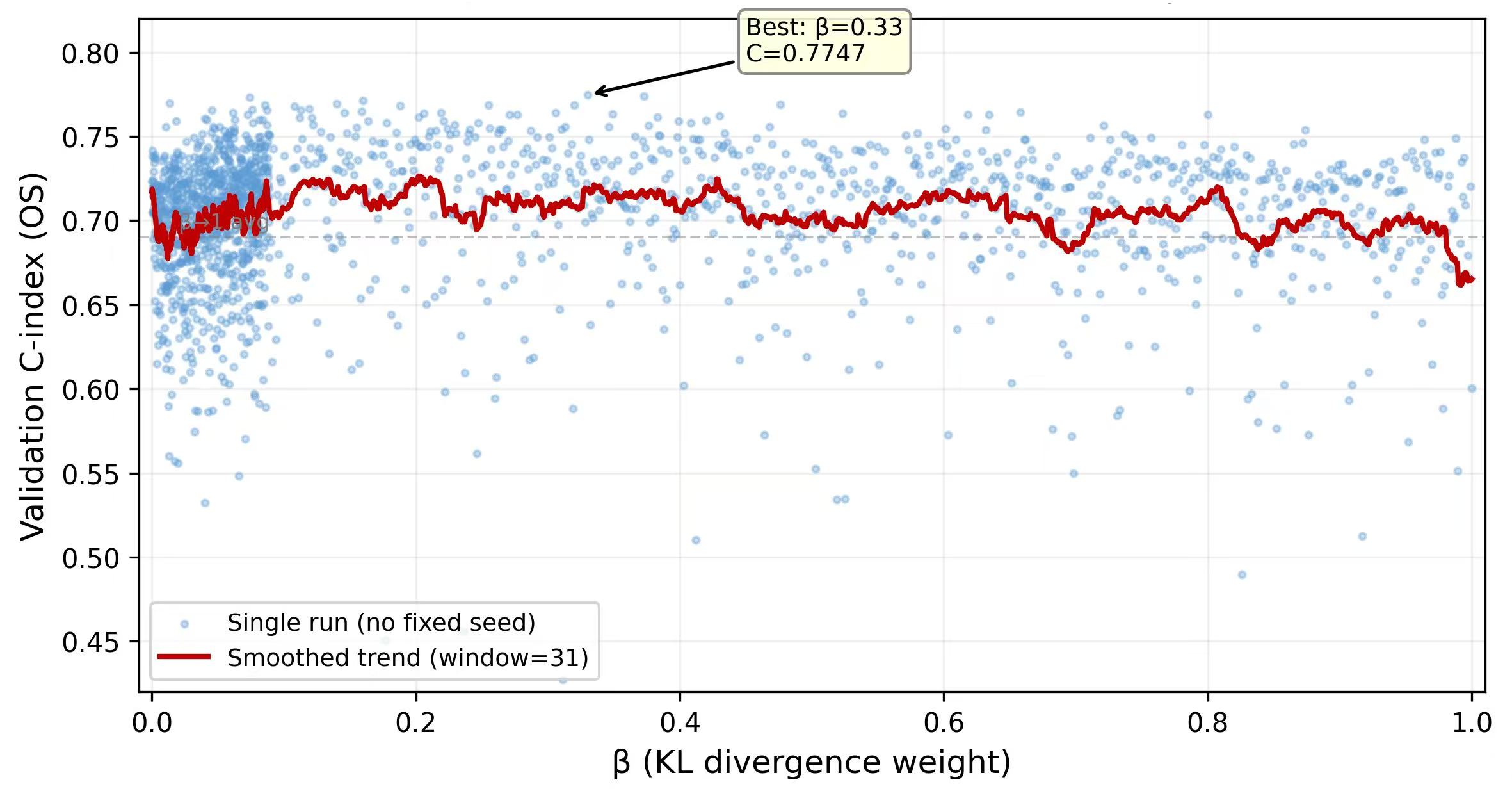}
\caption{Effect of KL regularization weight $\beta$ on validation C-index.}
\label{fig:beta_sweep}
\end{figure}

\noindent\textbf{Design Axis I: Effect of Latent Regularization Scale}.
We first examine how the scale of latent regularization influences survival
performance under Cox supervision. Figure~\ref{fig:beta_sweep} shows validation
C-index as a function of the KL weight $\beta$. Performance exhibits a clear
peak around $\beta \approx 0.33$, while both stronger ($\beta=1.0$) and weaker
regularization degrade survival discrimination.

Analysis of training loss magnitudes reveals that the Cox objective dominates
optimization, whereas KL-, MMD-, and HSIC-based regularizers operate at
substantially smaller scales. This imbalance explains why survival performance
is primarily governed by the effective strength of latent regularization rather
than the specific divergence formulation.

\noindent\textbf{Design Axis II: Posterior Geometry and Dependence Control}.
We next evaluate whether alternative posterior geometries or dependence-aware
regularization provide additional benefit when regularization scale is held
constant. Table~\ref{tab:main_best} reports the best single-run validation
C-index achieved by KL, $\beta$-KL, MMD, and MMD+HSIC variants under identical
architectures and training protocols.

\begin{table}[H]
\centering
\caption{Best validation C-index (single-run) under identical training settings.}
\label{tab:main_best}
\begin{tabular}{lc}
\toprule
Model & Best C-index \\
\midrule
KL ($\beta=1.0$) & 0.7138 \\
$\beta$-KL ($\beta=0.33$) & 0.7419 \\
MMD & 0.7121 \\
MMD+HSIC & 0.7140 \\
\bottomrule
\end{tabular}
\end{table}

\begin{figure}[H]
\centering
\includegraphics[width=0.85\linewidth]{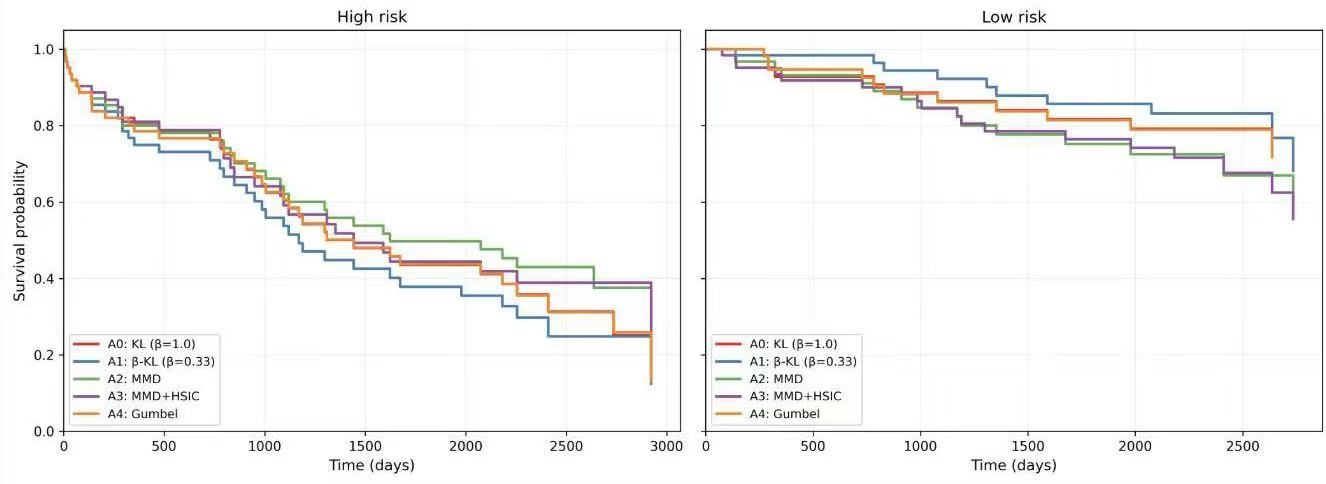}
\caption{Kaplan--Meier survival curves on the validation set (median risk split).}
\label{fig:km_gumbel}
\end{figure}

Despite modifying posterior geometry or suppressing residual-input dependence,
MMD- and HSIC-based variants fail to consistently outperform appropriately
scaled $\beta$-KL regularization. These results suggest that, under
survival-driven optimization, posterior geometry alone offers limited benefit
without careful calibration of regularization magnitude.

\noindent\textbf{Design Axis III: Latent Space Structure and Risk Alignment}.
We then investigate whether explicitly structuring the latent space improves
alignment between learned representations and survival risk. A hybrid
continuous--discrete latent formulation based on Gumbel--Softmax is evaluated
under the same backbone and training protocol.
The Gumbel--Softmax model achieves the highest validation C-index of 0.7788,
exceeding the best continuous baseline ($\beta$-KL, $\beta=0.33$) by +0.0369.
Kaplan--Meier analysis on the validation cohort further confirms its
discriminative ability (Fig.~\ref{fig:km_gumbel}), yielding a hazard ratio of 5.95
under a median risk split (log-rank $p = 2.1\times10^{-5}$).

\begin{figure}[H]
\centering
\includegraphics[width=0.85\linewidth]{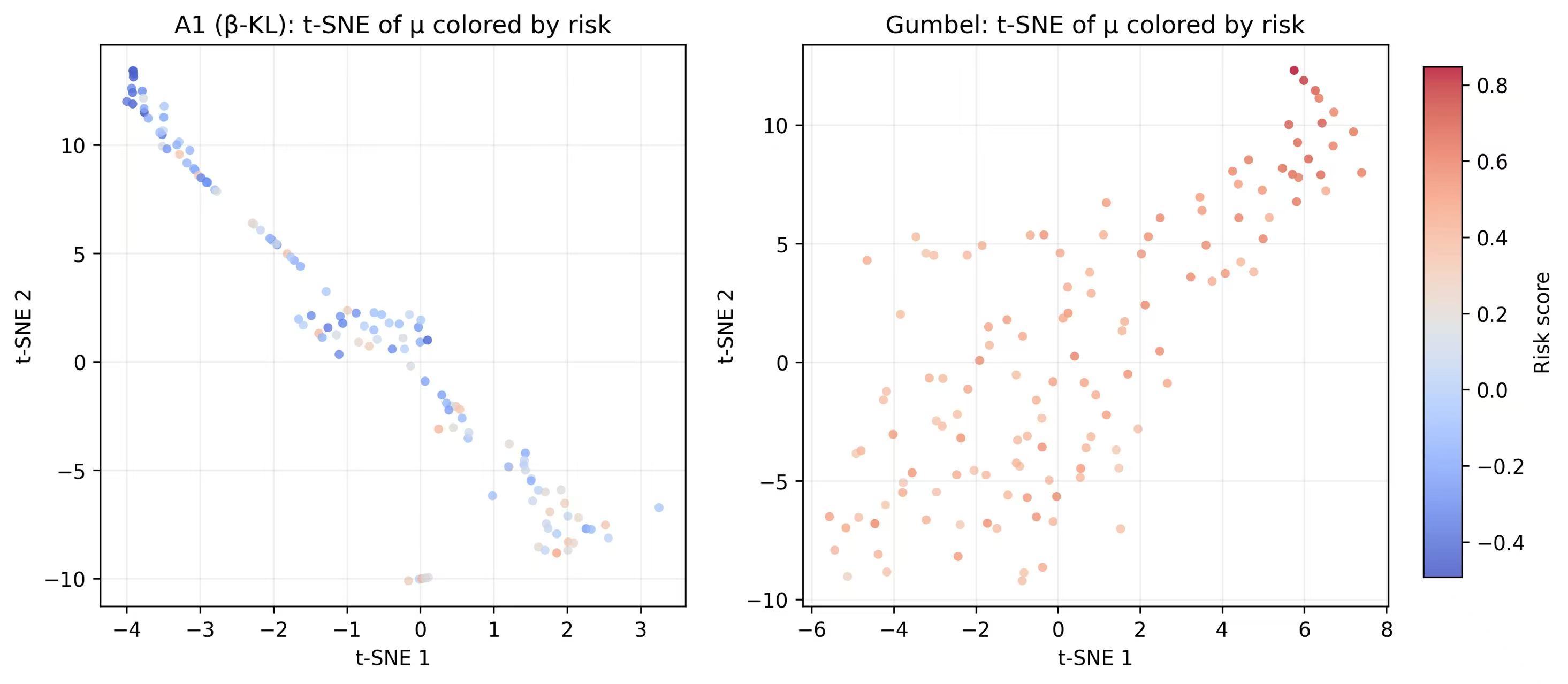}
\caption{t-SNE visualization of continuous latent representations colored by
predicted survival risk.}
\label{fig:latent_risk_tsne}
\end{figure}

To understand the source of this improvement, we analyze the alignment between
continuous latent representations and predicted survival risk.
Figure~\ref{fig:latent_risk_tsne} visualizes the latent mean $\mu$ colored by
predicted risk under identical t-SNE settings. Compared to the $\beta$-KL
baseline, the hybrid model exhibits a smoother and more monotonic risk gradient,
indicating improved global ordering in latent space.

Quantitatively, the Spearman correlation between the first principal component of
$\mu$ and predicted risk increases from $|\rho|=0.516$ to $|\rho|=0.637$, while
the mean absolute correlation across latent dimensions increases from 0.368 to
0.417, confirming stronger global alignment.



\noindent\textbf{MO-RiskVAE as a Principled Instantiation}.
Taken together, these results demonstrate that robust multimodal survival
modeling is primarily governed by appropriately scaled latent regularization
and latent structures that align with survival risk gradients. Based on these
findings, we instantiate MO-RiskVAE by combining moderately relaxed KL
regularization with structured latent representations, while avoiding
additional regularizers that do not yield consistent benefit. This
instantiation reflects a principled synthesis of the design axes validated
above, rather than an independent architectural proposal.

\section{Conclusion}

We systematically investigated latent modeling strategies for multimodal
survival prediction in multiple myeloma within a controlled extension of the
MyeVAE framework. By isolating regularization scale, posterior geometry, and
latent space structure under identical architectures and optimization
protocols, we clarified how latent design choices interact with
survival-driven objectives.

Our results show that preserving prognostically relevant variation primarily
depends on appropriately relaxing posterior constraints relative to the
dominant Cox loss. While alternative regularization mechanisms such as
$\beta$-KL, MMD, and HSIC differ in form, their impact is largely governed by
effective regularization scale, rendering divergence choice alone insufficient
without calibration.

We further demonstrate that explicit latent space structuring can improve
survival modeling beyond regularization alone. A hybrid continuous--discrete
latent formulation based on Gumbel--Softmax enhances global alignment between
continuous latent representations and survival risk gradients, despite the
absence of stable discrete subtype discovery under survival supervision. This
indicates that the benefit of hybrid latent design in moderate-size cohorts
arises mainly from improved organization of continuous risk-related variation.

Based on these findings, we instantiate \textbf{MO-RiskVAE} as a principled
multimodal survival model that combines moderately relaxed latent
regularization with structured latent representations while avoiding
unnecessary complexity. Overall, our work emphasizes that robust multimodal
prognostic modeling depends less on increasingly sophisticated divergence
formulations than on latent representations whose scale and structure are
commensurate with survival-driven optimization.
%
%
%
%

\end{document}